\documentclass[runningheads]{llncs}
\usepackage{amsmath}
\usepackage{amssymb}
\usepackage{graphicx}
\usepackage{wrapfig}
\usepackage[linesnumbered,ruled]{algorithm2e}
\usepackage[noend]{algorithmic}
\usepackage{floatrow, caption}

\usepackage{tikz}
\usepackage{tikz-dependency}
\usetikzlibrary{tikzmark}

\usepackage{todonotes}

\newcommand{\Sch}{\hbox{$\mathbb{G}$}} 
\newcommand{\db}{\hbox{$\mathbb{D}$}} 
\newcommand{\ltype}{\textsf{lexT}} 
\newcommand{\dbtype}{\textsf{dbT}} 
\newcommand{\predE}{\textsf{getPred}} 
\newcommand{\auxSt}{\textsf{Ontology-Mappings}} 
\newcommand{\buildAtom}{\textsf{bAt}} 
\newcommand{\buildAtomOP}{\textsf{bAtOp}} 

\def\ie{\textit{i\,.e\,.}}
\def\eg{\textit{e\,.g\,.}}
\def\wrt{\textit{w\,.r\,.t\,.}}

\begin{document}
%
\title{Natural Language Querying System through Entity Enrichment}
%
\author{Joshua Amavi\inst{1} \and Mirian Halfeld-Ferrari\inst{2} \and Nicolas Hiot\inst{1, 2}}
\authorrunning{Amavi, Halfeld-Ferrari,  Hiot}
%
\institute{EnnovLabs, Paris, France  (\url{https://en.ennov.com/}) \\
    \and Universit\'e d'Orl\'eans, INSA CVL, LIFO EA, Orl\'eans, France \\
    \email{\{jamavi,nhiot\}@ennov.com~~ \{mirian,nicolas.hiot\}@univ-orleans.fr}}
\maketitle
\begin{abstract}
This paper focuses on a domain expert querying system over  databases. It presents a solution
designed for a French enterprise interested in offering a natural language interface for its clients.
The approach, based on entity enrichment,  aims at translating  natural language queries into database queries.
In this paper, the database is treated through a logical paradigm, suggesting the adaptability of our approach to different database models.
The  good precision of our method is shown through some  preliminary experiments. 


    \keywords{NLI \and NLP \and Database query  \and Question answering.}
\end{abstract}
\section{Introduction}
\label{sec:Intro}
Graph database querying systems adapted to  \textit{domain experts},  and not only to database experts, deserve great attention nowadays and become an important research topic. 
Query language such as SPARQL or CYPHER are powerful tools but require  knowledge of the  database structure  in order to retrieve information.
To simplify the accessibility of such databases, the  research of natural language interface (NLI) to (structured) databases receives considerable  attention today~\cite{wangCrossdomainNaturalLanguage2019,zhengQuestionAnsweringKnowledge2018}. The idea of NLI is to allow users to focus on the semantic of what they want rather than on how to retrieve it.

This paper describes a practical solution for simple natural language queries on an RDF database, developed for clients of Ennov, a French enterprise specialised in  building software solutions to the  life sciences industry.
Our proposal focuses on the enterprise needs, \ie,  factoid queries concerning instances of one RDF 'class', but achieves good results allowing  
(i) to envisage its use to other domains and (ii) to extend its ideas to more complex queries.
The proposal consists in translating a given natural language query (denoted as NL-query)  in a database query (denoted as DB-query). 
In this paper,  we use a logical formalism to express database and DB-queries which can be easily translated to any graph,  or relational model  (and thus to queries on SQL, SPARQL, etc).
 
In this context, the main contributions of this paper are:
\begin{itemize}
\item An original method, based on entities' enrichment,  for translating a NL-query into a DB-query. Indeed, entity extraction is a subtask of NLP (Natural Language Processing) and consists of identifying part of an unstructured text that represents a named entity. After identifying entities connected to a specific domain, a classification into different entity types is possible. Following this classification, some of them are merged and a  set of enriched entities is obtained. DB-queries are built from this set of enriched entities.
 
\item An approach composed by two distinct phases: a domain specific pre-proces\-sing step and a general query-generating step. The pre-processing step puts in place the general environment which guides query translation: lexicons are built (partially) from the information stored in the database, grammars and ontology mappings are set up.
Query-generating algorithms classify and enrich extracted entities and then transform the obtained set of  enriched entities into database queries.

\item A good-precision querying system. Our approach focus on restricted and specialized domain queries which imply a relatively small vocabulary (mostly composed by people  and technical terms appearing in the database instance). 
Our method takes advantages of this context and  gives  priority  to the use of grammar- and lexicon-  based tools.
The result is an efficient  and precise query translation system.

\item An approach  proposing  a non disambiguation of the natural language queries. 
Indeed, instead of resolving the  ambiguity problem intrinsic to natural language, we adopt a lazy approach and consider all possible interpretations, generating all possible database queries.
This option avoids the expensive disambiguation process  and speed up the whole performance.
The same idea is used to solve ambiguity coming from the use of coordinating conjunctions.

\end{itemize}

Our querying system is available over an RDF database storing  information about medical documents.
The system  translates a NL-query into a  DB-query offering a user-friendly interface.
Let us briefly introduce each of these queries together with a running example.

 \paragraph{\textbf{NL-query.}}
\textsf{Ennov}'s motivation is to offer  a querying system capable to allow its users
to perform the so-called \textit{facet search},   narrowing down search results by applying multiple filters. 
Accepted queries are those  requiring information on instances of one unique RDF class (denoted here as solar-class).
An allowed query  selects only the solar-class instances (the nodes of the given type) via properties 
 having the solar-class as domain or range (the in- or out-edges).
For instance, supposing that \textit{Book}  is a solar-class, a query requiring book instances edited by doctor Alice on cardiology  after year 2018  is an allowed query.
On the contrary, a query  requiring book instances edited by doctor Alice who is cardiologist, is not allowed since \textit{is cardiologist} is not a property (and edge) of the current solar-class.
If the user wants to identify doctors who are writers, then he has, firstly, to change  his solar-class specification.
In other terms, our query are simple queries 
identifying instances of one class (even if it can also renders values concerning properties of these instances). The  NL-query follows the format \textit{Find books which...} (establishing Book as the solar-class), \textit{Find doctors who...}
(Doctors as the solar-class), etc.

\vspace{0.05cm}
Now, as a running example, we introduce query $Q_{run}$.  We use it in the rest of the paper  to show, step by step, how to obtain  a DB-query. When talking specifically about the NL-query version we can write  $NLQ_{run}$. 

\vspace{-0.1cm}
\small

\begin{quote}
\textit{Find books with title 'Principles of Medicine' co-authored by Bob and Alice and whose price is less than 30 dollars.} 
\end{quote}

\normalsize

 \paragraph{\textbf{DB-query.}}
We use a logical paradigm to express RDF databases  and queries.
We write \textit{Book(Anatomy)}  
to express that $Anatomy$ is an instance of class $Book$  and
\textit{writtenBy(Anatomy, Bob)} to express that $Anatomy$ has value $Bob$ for property $writtenBy$.

 We briefly introduce this logic formalism  (refer to~\cite{AHV95} for some background on this aspect).
 An  {\em atom} has  one of the forms:
 ($i$) $P(t_1, . . . , t_n)$, where $P$ is an $n$-ary predicate and $ t_1 , . . . , t_n$ are terms (terms are constants or variables);
($ii$) $\top$ (true) or $\bot$ (false);
 ($iii$) $(t_1 ~op~  \alpha_2)$, where 
 $t_1$ is a  term,  
 $\alpha_2$ is a term
 or    a character string,  
 and $op$ is  a comparison operator.
 A \textit{fact} is  an  atom  $P(u)$ where $ u$ has only constants.  
A \textit{database schema} is a set of predicates \Sch\ and a \textit{database instance} is a set of facts (denoted by \db) on \Sch. 

A \textit{(conjunctive) query}  $q$  over  a given schema has the rule-form
$R_0 (u_0) \leftarrow R_1(u_1) \dots  R_n(u_n), comp_1, \dots, comp_m$
where $n \geq 0$, $R_i$ ($0 \leq i \leq n$) are predicate names,  $u_i$ are   tuples of terms of
appropriate arity and $comp_j$  ($0 \leq j \leq m$) are comparison formulas involving variables appearing in at least one  tuple from $u_1$ to $u_n$.
We denote $head(q)$ (respect. $body(q)$)  the expression on the left hand-side (respect. right hand-side) of $q$ .
The answers for  $q$ are tuples $t$ only with constants.
For each $t$ there exists a mapping $h_t$ (which maps variables to constants and a constant to itself)
such that
$\{R_1(h_t((u_1)),  \dots,  R_n(h_t((u_n))\} \subseteq \db$,  the conjunction of all $h_t(comp_j)$ is evaluated to true
(according to the usual semantic of opertors $op$) and $h_t(u_0)= t$.
In this rule-based formalism, the union is expressed by allowing more than one rule with the same head.
For instance, 
$q(X) \leftarrow writtenBy(X, Bob)$ together with
$q(X) \leftarrow editedBy(X, Bob)$
express a query looking for documents written or edited by \textit{Bob}.

\vspace{0.05cm}
\textit{Book} is  the solar-class in  $NLQ_{run}$ and  thus the DB-query should return the identifiers of book instances. 
 The following conjunctive DB-query 
 is the $DBQ_{run}$ -- it includes all the  conditions imposed on the books being selected.
 
 \vspace{0.1cm}
\small
\begin{tabular}{ll}
$ Q(x)  \leftarrow$ & $ Book (x), hasTitle (x, y_1), writtenBy (x, y_2), Person (y_2), writtenBy (x, y_3) , $\\
& $ Person(y_3), hasPrice (x, y_4),$ $(y_1 = 'Principles ~of ~Medicine'),$\\
& $ (y_2= \text{:bob}), (y_3 = \text{:alice}), (y_4 < 30)$. \hfill$\Box$
\end{tabular}
\normalsize

\paragraph{\textbf{Paper Organization.}} 
 Our approach is depicted in Sections~\ref{sec:Eext} and~\ref{sec:E2Q} while implementation and testing results are presented in Section~\ref{sec:implementation}.
 Section~\ref{sec:rw} concludes the paper with some related work and perspectives.

\section{Entity Extraction and Enrichment}
\label{sec:Eext}

In this paper we define a \textit{simple entity} as the tuple
$E = (\mathcal{V}, \mathcal{T},  m)$ where
$\mathcal{V}$  and $\mathcal{T}$ are lists containing values and lexical types, respectively,  
and $m$ is a mapping such that  $\forall v \in \mathcal{V}, \exists T \subseteq \mathcal{T}, m(v) \rightarrow T$.

Indeed,  during the entity extraction phase, ambiguity, an inherent problem in many steps of natural language processing,
exists: it concerns  the type (\eg \textit{Paris} can refer to a city or a person) or the value (\eg, several people in the database have the same name).
Generally, we seek to eliminate this ambiguity by keeping only the most likely solution.
Such a solution may introduce contradiction \wrt\ the text semantics.
In our approach we explicitly reveal ambiguity (a value may be associated to different types)  and we 
 keep track of multiple interpretations during this extraction step, a convenient solution
for querying, if we consider the situations where ambiguity can be represented by an  \textit{OR} connective.

Entity Extraction (EE) or Named Entity Recognition (NER) is a subtask of NLP and consists of identifying part of an unstructured text that represents a named entity.
This task can be performed either by grammar-based techniques  or by a statistical models such machine learning (refer to~\cite{JM19} for a complete introduction in the domain).
Statistical approaches are widely used in the industry because they offer good results with the latest research and the work of giants like Google, Facebook or IBM.
However, these approaches mainly require a lot of data to get good results, implying high costs.
More conventional  grammar-based methods are very useful for dealing with small data sets.

\vspace{0.1cm}
\noindent
\textbf{Entity extraction.} Our proposal consists in applying different grammar- or lexicon-based 
methods together in order to extract simple entities from a given
NL-query. The combination of their results allow us to improve entity extraction.
The initial parsing step is followed by two different entity extraction methods.
One consists of looking up on dictionaries (lexicons) for qualifying an entity. 
The other is based on  local grammars.
Notice that the dependency tree resulting from the parsing phase may be used for guiding entity extraction with local grammars.

\begin{figure}
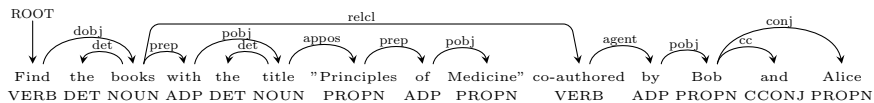

  \tiny
  \centering
  \begin{dependency}[theme=simple, edge vertical padding=1ex]
    \begin{deptext}[row sep=.5ex]
      Find \& the \& books \& with \& the \& title \& "Principles \& of \& Medicine" \& co-authored \& by \& Bob \& and \& Alice \\
      VERB \& DET \& NOUN \& ADP \& DET \& NOUN \& PROPN \& ADP \& PROPN \& VERB \& ADP \& PROPN \& CCONJ \& PROPN  \\
    \end{deptext}

    \deproot[edge height=10ex]{1}{ROOT}
    \depedge{3}{2}{det}
    \depedge{1}{3}{dobj}
    \depedge{3}{4}{prep}
    \depedge{6}{5}{det}
    \depedge{4}{6}{pobj}
    \depedge{6}{7}{appos}
    \depedge{7}{8}{prep}
    \depedge{8}{9}{pobj}
    \depedge[segmented edge, edge height=7ex]{3}{10}{relcl}
    \depedge{10}{11}{agent}
    \depedge{11}{12}{pobj}
    \depedge{12}{13}{cc}
    \depedge{12}{14}{conj}
  \end{dependency}
  \caption{Dependency tree and POS tagging}
  \label{fig:depTree}
\end{figure}

\vspace{0.1cm}
\noindent
\textit{Parsing.}
In our approach, 
tokenization (\ie, determine the words and punctuation), 
part-of-speech (POS) tagging (\ie, determine the role of the word in a sentence), 
lemmatization (\ie,  determine  word canonical form), 
stematization (\ie, strip word suffixes and prefixes) and dependence analysis are achieved by 
\textsf{SpaCy}~\cite{spacy} built on
a convolutional neural network (CNN) \cite{honnibalSpacyNaturalLanguage2017} learned from a generic English corpus~\cite{weischedelOntoNotesLargeTraining2011}.
The dependency tree produced by \textsf{SpaCy} \cite{honnibal-johnson} guides different choices in some of the following steps of our approach.
Fig~\ref{fig:depTree} illustrates part of the dependency tree for $NLQ_{run}$.

\vspace{-1em}
\vspace{-1.5em}

%


\addtolength{\textfloatsep}{-2em}
\begin{figure}[t]
  \scriptsize
  \begin{tabular}{l|l c |l|l|} \cline{4-5}
    \multicolumn{2}{c}{\textbf{LEXICONS}} & \hspace{.8cm}                                & \multicolumn{2}{c|}{\textbf{INVERSE INDEX}}                                              \\ \cline{4-5}
    \textit{EntityName}                   & \textit{Lexemes}                             &                                             & Pointer to & Index  Entry (word from text) \\ \cline{1-2} \cline{4-5}
    :alice                                & \textit{Alice, Wonderful, Wonderful Alice}   &                                             & :alice     & \textit{Alice}                \\
    :bob                                  & \textit{Sponge, Bob, Sponge Bob}             &                                             & :alice     & \textit{Alice Wonderful}      \\
    $\dots$                               &                                              &                                             & $\dots$    &                               \\
    author                                & \textit{co-authored, written by, created by} &                                             & :bob       & \textit{Bob}                  \\
    $\dots$                               &                                              &                                             & author     & \textit{co-authored}          \\
    lt                                    & \textit{less than}                           &                                             & lt         & \textit{less than}            \\
    $\dots$                               &                                              &                                             & :alice     & \textit{Wonderful}            \\
    $\dots$                               &                                              &                                             & $\dots$    &                               \\ \cline{1-2} \cline{4-5}
  \end{tabular}

  \vspace{0.3cm}

  \begin{tabular}{|l|l|l|c|l|l|cl|l}
    \multicolumn{7}{c}{\auxSt} & \multicolumn{2}{c}{\textsf{Operator Dictionary}}                                                                                                                    \\ \cline{1-3} \cline{5-6}
    \textit{EntityName}        & LexType                                          & DBType     & \hspace{0.3cm} & DBType     & Predicate         & \hspace{0.4cm} & \textit{EntityName} & Comparator \\
    \cline{1-3} \cline{5-6}  \cline{8-9}
    :alice                     & Person                                           & Person     &                & Person     & Person (X)        &                & lt                  & $<$        \\
    author                     & Context                                          & Author     &                & Author     & writtenBy (\_, X) &                & leq                 & $\leq$     \\
    :bob                       & Person                                           & Person     &                & Book       & Book (X)          &                & gt                  & $>$        \\
    title                      & Context                                          & Title      &                & Title      & hasTitle (\_, X)  &                & geq                 & $\geq$     \\
    :alice                     & Doctor                                           & MedicalDoc &                & MedicalDoc & Doctor (X)        &                & \dots               & \dots      \\

    \cline{1-3} \cline{5-6}
  \end{tabular}

  \caption{Auxiliary structures built in the pre-processing phase}
  \label{fig:Overview1}
\end{figure}

\noindent
\textit{Lexicons.} When, as in our case, we  deal  with entity extraction founded on a relatively small database, 
 one could envisage to  verify, for each 
 value in the database, whether it appears in the text.
However, to improve performance and ensure low coupling, our option consists in building  lexicons (\ie, lookup tables) from the database and  then in using them as
dictionaries containing a summary of the database instance.

Fig.~\ref{fig:Overview1}  illustrates an inverse index pointing to lexicons.
It helps finding \textit{EntityValue} quickly.
For instance, if \textit{Wonderful Alice} is the string found in the text, the index allows us to find how to refer to it, \ie, with its \textit{EntityValue}, \text{:alice}.
With \textit{EntityValue} we have access to information stored in  auxiliary structures  denoted here by \auxSt.
In order to simplify notations,  we consider the existence of three functions that render information over \auxSt. They are:
(1) \ltype: given \textit{EntityValue}, the function renders its lexical type indicating entity's lexical role, detected during the lexical analysis;
(2)  \dbtype: given \textit{EntityValue}, the function renders its database type which indicates the semantic associated to the entity in our RDF database and
(3) \predE: given the database type, the function renders  the associated predicate  which is the one  we should use in the DB-query. Moreover, the  arity of predicates   is indicated together with the argument position reserved for an entity having this database type.
According to the example in Fig.~\ref{fig:Overview1}, we have:
$\ltype(:alice) = Person$, 
$\dbtype(:alice) = Person$
and 
$\predE(Person) = Person (X_{person})$.  
Similarly,  lexeme \textit{co-authored by} refers to the entity value \textit{author} and we obtain
$\ltype(author) = Context$, 
$\dbtype(author) = Author$
and 
$\predE(Author) = writtenBy(X, Y_{author})$.

\begin{wrapfigure}[9]{R}{0.4 \textwidth}
  \vspace{-2.3em}
  \centering
  \scriptsize
  \begin{tabular}{|l|c|c|} \hline
    Entity & \textit{EntityValue} & \ltype   \\ \hline
    $E_1$  & book                 & Context  \\
    $E_2$  & title                & Context  \\
    $E_3$  & author               & Context  \\
    $E_4$  & :bob                 & Person   \\
    $E_5$  & :alice               & Person   \\
    $E_6$  & price                & Context  \\
    $E_7$  & lt                   & Operator \\ \hline
    $E_8$  & 'Princ. of Medecine' & Text     \\
    $E_9$  & 30                   & Number   \\ \hline
  \end{tabular}
  \caption{Simple entities extracted from $NLQ_{run}$}
  \label{fig:Ent-V1}
  \vspace{-1em}
\end{wrapfigure}

\vspace{0.1cm}
\noindent
\textit{Local grammar.}
There is no sense in storing possible values for attributes associated to huge domains.
Dates, general numerical attributes (\eg, prices, weight, etc) or even publication or section titles are not stored in our database.
Entity extraction, for such cases, is based on local grammars. 
Currently we have designed 15 local grammars as a support for our entity extraction mechanism.

\begin{example}
\label{ex:simpleEnt}
We show our $NLQ_{run}$ with expressions in boxes  indicating the entities detected after the extraction step.
Fig.~\ref{fig:Ent-V1} summarizes obtained entities: $E_1$-$E_7$ via Lexicons while $E_8$-$E_9$
are detected  by local grammars.
Fig.~\ref{fig:Ent-V1} does not represent set notation to avoid figure overload.
Here, each entity has only one value and each value is associate to a singleton.

\small
\begin{quote}
Find the  \fbox{books\tiny{1}}  with  the \fbox{title\tiny{2}} \fbox{"Principles  of  Medicine"\tiny{8}}  \fbox{co-authored  by\tiny{3}}  \fbox{Bob\tiny{4}}  and  \fbox{Alice\tiny{5}}
  and  whose  \fbox{price\tiny{6}}  is  \fbox{less  than\tiny{7}}  \fbox{30\tiny{9}}  dollars. $\hfill\Box$
  \end{quote}
\normalsize

 \normalsize
\end{example}

Once we have extracted simple entities, we classify them into categories  in order to  decide about their fusion or not.
Our goal is to build enriched entities which concentrate information initially available in the detected simple entities.
An enriched entity  is a first-class citizen which will guide the construction of DB-queries while simple entities are auxiliary ones considered as second-class citizens and committed to integrate  the enriched entity.

\vspace{0.1cm}
\noindent
\textbf{Entity Enrichment.}
An \textit{enriched entity}  is a relation. It can thus  be seen as table (as in the relational model)
with schema  $E_e [EntityValue, DBType, $ $LexType, op]$.
The entity $E_e$  itself is a relation instance, \ie,  a set of functions (tuples) associating each attribute in the schema to a value.
Thus, each tuple maps:
(i)  $EntityValue$ to the value $v$  of an entity as represented in a Lexicon (Fig~\ref{fig:Ent-V1}),
(ii) $DBType$ and  $LexType$ to \dbtype($v$) and \ltype($v$), respectively (Fig~\ref{fig:Overview1}),
and 
(iii) $op$ to a comparison operation indicating the kind of comparison   imposed on the entity value. 
By default, the comparison  operation $op$ is \textit{equal to}.

On the other hand, we distinguish the following classes of simple entities:

\noindent
$\bullet$ A \textit{reference entity} is the one chosen to evolve, \ie, to be transformed into an enriched entity.
It corresponds to an instance value in a database (\eg, people names, document titles, dates, etc) .

\noindent
$\bullet$ An \textit{operator entity} $E_{cp}$ represents words which indicate that another (reference) entity $E=(\mathcal{V}, \mathcal{T}, m)$ is constrained by 
a comparison condition.
A connection between $E$ and $E_{cp}$ in  the dependency tree  determines $E$ as the reference entity. 
Then, $E$ evolves to an enriched entity  $E_e$ such that for each $v \in \mathcal{V}$
we have 
$(v, \dbtype (v), \ltype(v), op)$ in $E_e$ where $op$ is defined by $E_{cp}$, according to an available dictionary (see example in Fig.~\ref{fig:Overview1}).
Notice also that $op$  corresponds to the operator used in the DB-query (Section~\ref{sec:Intro}) where  \textit{comp} atoms 
have the general format $(t_1 ~op~ \alpha_2)$.
In Example~\ref{ex:simpleEnt}, expression \textit{less than} is an operator entity having $30$ as its reference entity.

\noindent
$\bullet$ A \textit{context entity} $E_C $ gives information about the type of another (reference) entity.
Once again, the dependency tree obtained during the parsing, determines the reference entity $E$ which evolves to a new enriched entity $E_e$ according to Algorithm~\ref{algo:small}. 
In Example~\ref{ex:simpleEnt}, \textit{price} is a context entity and $30$ as its reference.
Similarly, \textit{title} is  a context entity and \textit{Principles of Medicine} its reference~(Fig~\ref{fig:depTree}).

We remark that the first For-loop of Algorithm~\ref{algo:small} transforms a simple entity into an enriched one.
Given a simple entity $E$, $extend(E)$ is the relation instance obtained by converting $E$ into its extended counterpart.
Notice that the entity evolution process starts with a simple entity which becomes an enriched entity, but such an enriched entity can continue evolving.


\begin{example}
\label{ex-enrEt}
From Fig.~\ref{fig:Ent-V1} and our auxiliary structures (partially depicted in Fig.~\ref{fig:Overview1}) we obtain the following enriched entities:\\
\footnotesize
$E_{e0}= \{( book, Book, Context, =), \}$; \\
$E_{e1}= \{('Princ.~of~Medicine', Text, Text, =), $ \\
$~~~~~~~~~~~('Princ.~of~Medicine', Title, Context, =)\}$\\
$E_{e2} = \{(\text{:bob}, Person, Person, =), (\text{:bob}, Author, Context, =)\}$ \\
$E_{e3} = \{(\text{:alice}, Person, Person, =), (\text{:alice}, Author, Context, =)\}$\\
$E_{e4} = \{(30, Number, Number, <),  (30, Price, Context, <)\}$.\\

\normalsize
All entities are enriched ones.
$E_{e1}$ results from the integration of context entity $E_2$ into $E_8$,
$E_{e2}$  results from  integration of $E_3$ into $E_4$ and 
$E_{e3}$ results from integration of  $E_3$ into $E_5$.
Notice that coordinating conjunction \textit{and} in $NLQ_{run}$ 
implies the existence of these two latter independent enriched entities.
$E_{e4}$ results from the integration of operator entity $E_7$ to $E_9$.
Entity $E_{e0}$ is just the enriched version of $E_1$. It corresponds to the solar-class.~\hfill$\Box$

\end{example}

\begin{algorithm}[t]
  \footnotesize
  \begin{algorithmic}[1]
    \REQUIRE
    $E_C = (\mathcal{V}_C, \mathcal{T}_C, m_C)$  and  $E=(\mathcal{V}, \mathcal{T},  m)$ 

    \ENSURE
    $E_e$: an instance over schema $E_e [EntityValue, DBType, $ $LexType, op]$

    \FORALL{$v \in \mathcal{V} $}
      \STATE Insert $(v, \dbtype (v) , \ltype(v), op)$ in $E_e$;
      \FORALL{$u \in  \mathcal{V}_C$}
        \FORALL{$v \in \mathcal{V}$}
          \STATE Insert $(v,\dbtype(u) , \ltype(u), op)$ in $E_e$;
        \ENDFOR
      \ENDFOR
    \ENDFOR
  \end{algorithmic}

  \caption{contextEnrichment}
  \label{algo:small}
\end{algorithm}

Now, a  NL-query may include multiple conditions (or filters) connected  by coordinating conjunctions.
Our current version deals only with  \textit{and} and \textit{or}, even if we intend to extend this initial proposal to more complex coordinating conjunctions such as \textit{nor}, \textit{for}, etc\,.
Coordinating conjunctions are  expressed through logical formulas which
guide the construction of the DB-query, by specifying: (i) the kind of atoms \textit{comp} it will have and (ii) whether the query is defined by one of several rules.
Taking into account coordinating conjunctions implies entity enrichment.
Let $E_1= (\mathcal{T}_{1}, \mathcal{V}_{1} , m_1)$ and $E_2= (\mathcal{T}_{2}, \mathcal{V}_{2} , m_2)$ be two simple entities having the same LexType.
If, in  the  query text,  these two entities are connected by an \textit{or}, they are merged, forming a new enriched entity 
composed by $extend(E_1) \cup extend(E_2)$.
 The original entities do not exist any more.
Otherwise, if  in  the text,  these two entities are connected by an \textit{and}, they are kept as independent ones.

\begin{example}
\label{ex:or}
In Example~\ref{ex-enrEt}, 
$E_{e2} = \{(\text{:bob}, Person, Person, =), (\text{:bob}, Author, $ \linebreak $Context, =)\}$ and
$E_{e3} = \{(\text{:alice}, Person, Person, =), (\text{:alice}, Author, Context,$ $=)\}$ are independent entities.
When considering $NLQ_{run}$ these entities do not merge because the coordinating conjunction is an \textit{and}.
If  we  change the sentence to  '\textit{written by Bob or Alice}', entities are merged resulting in:

\vspace{.5em}
$E_{enew} = \{(\text{:bob}, Person, Person, =), (\text{:bob}, Author, $ $Context, =),$\\
$~~~~~~~~~~~~~~~~~~(\text{:alice}, Person, Person, =), (\text{:alice}, Author, Context, =)\}$.~\hfill$\Box$
\end{example}

Queries may have multiple coordinating conjunctions as illustrate in sentence   \textit{written by Alice or Bob and Charlie} and, in this case, 
its interpretation (due to the ambiguity of natural language) can vary, 
resulting in the logic formula  $\mathcal{F}_1 \equiv$ $(X= \text{:alice})  \lor ((X= \text{:bob}) \land  (X= \text{:charlie}))$ or
in the formula  $\mathcal{F}_2 \equiv$ $((X= \text{:alice}) \lor (X= \text{:bob})) \land (X= \text{:charlie})$. 
 To avoid erroneous query answers, one may envisage to take into account all the alternative interpretations, or to give choices to the user.
In our approach we do not plan interactions with the user and thus, 
we propose to consider a larger interpretation, \ie, to overcome ambiguity by 
replacing a mixed \textit{and}-\textit{or} sentence by an only-\textit{or} sentence.
Thus, let   $\mathcal{F}$  be the logic formula obtained from a sentence with coordinate conjunctions.
If $\mathcal{F}$ contains both $\lor$ and $\land$, then we replace it by the formula $\mathcal{F}'$ composed only of $\lor$.
The idea is based on the fact that any answer satisfying  $\mathcal{F}$ also satisfies $\mathcal{F}'$.
In that way, when multiple coordinate conjunctions are present, the DB-query will be represented by multiple rules with the same head.
\vspace{-1em}

\section{Building DB-Queries from Enriched Entities}
\label{sec:E2Q}
\vspace{-.5em}
Once our NL-query is analysed and all enriched entities are completed, the DB-query is generated by
Algorithm~\ref{algo:EntitiesToQueries}.
The algorithm starts by considering entity $E_{e0}$ (line~\ref{algoL:solar}) which has a special role since it specifies the solar-class, \ie, the class on which the query  focuses.
The query is initialized with a body composed by one unique  atom over the predicate associated to the solar-class.
Notice the use of function $\buildAtom\ $ which is responsible for building an atom for the query being constructed. 
The predicate symbol to be used in the construction of an atom is found via the  value of attribute $DBType$ in $E_{e0}$ -- which is then used as an input for function \predE.
In Example~\ref{ex-enrEt},   \textit{Book} is the value of attribute $DBType$ in $E_{e0}$ and the name of the associated unary predicate. Atom $A_0$ in our case is $Book(x)$.
Notice that Algorithm~\ref{algo:EntitiesToQueries} builds only queries whose answers are books' identifiers (\ie, instantiations of $x$). Our initial query is thus $q(x) \leftarrow Book( x)$.

Lines~\ref{algoL:startEC} to~\ref{algoL:endEC} of  Algorithm~\ref{algo:EntitiesToQueries} consider entities $E_e$  enriched with a context entity.
If in  $E_e$  there are more than one tuple $t$ for which the value of attribute $lexType$ is ''Context'', then $E_e$ 
is an entity obtained after taking into account coordinating conjunction \textit{or}. 
Each tuple $t$ having value ''Context'' for attribute $lexType$ has to be grouped together with the tuple $t'$ representing its reference. On line~\ref{algoL:stECinParts}, Algorithm~\ref{algo:EntitiesToQueries} groups  each $t$ with another $t' \in E_e$ having the same value for attribute $EntityValue$.
From information in $t$ and $t'$ we build a list $l$, added to set \textit{Parts}.
Each $l \in Parts$ is a list of atoms  to be added to the body of the query under construction.
Notice that  Algorithm~\ref{algo:EntitiesToQueries} divides $E_e$'s tuples into parts (or lists). 
Each list in $Parts$ generate a new distinct query with the same $head$. 
Indeed, on line~\ref{algoL:l2q}, in  the for-loop,  each list $l$  is used to create a new query $q'$ -- continuing the construction of a query $q$ already in $\mathcal{Q}$. If there are more than one list $l$ in $Parts$, there will be more than one query $q'$.

From Example~\ref{ex-enrEt}, $E_{e2}$ has two tuples.
Let $t_1$ be the first tuple for which $\predE(t_1(DBType)) = \predE(Person)) = Person$
and $t_2$ be the second one for which $\predE(t_2(DBType)) = \predE(Author)) = writtenBy $.
The function \buildAtom\ can be used to build atoms that will be added to $body(q)$.
Notice that \buildAtom\  also takes into account information concerning  positions marked as the place for  the entity value in the atom being built. Thus, the new variable $y$ representing the entity is placed accordingly. 
In binary predicates $x$ is always the other variable.
Atoms $comp$ may associate  a value to variable $y$. 
Thus, on line~\ref{algoL:stECinParts} list $\langle Person (y_2),$ $ writtenBy(x, y_2),$  $(y_2= \text{:bob})   \rangle$ is added to $Parts$
and on line~\ref{algoL:l2q}  the query being built is
$q(x) \leftarrow Book(x),$ $Person (y_2),$ $writtenBy(x, y_2),$  $(y_2= \text{:bob})$.
The result obtained with entity $E_{e3}$ is similar.
However, entities $E_{e1}$ and $E_{e4}$ are treated in a different way since their lexical types are \textit{Text} and \textit{Number}, respectively.
These entities are treated on lines~\ref{algoL:startNonEC}-\ref{algoL:endNonEC}.
For instance, $E_{e1}$ gives rise to list 
$\langle hasTitle(x,y_1),$ $ (y_1= "Princ. ~of ~Medicine") \rangle$.
After considering all entities, Algorithm~\ref{algo:EntitiesToQueries} returns set $\mathcal{Q}$ with  the following  DB-query:

\vspace{-0.5cm}
\small
\begin{equation*}
	\begin{split}
		q(x) \leftarrow & Book(x), hasTitle(x,y_1), Person(y_2),\\
		& writtenBy(x, y_2), Person(y_3), writtenBy(x, y_3), asPrice(x, y_4),\\
		& (y_1= "Princ. ~of ~Medicine"), (y_2 = \text{:bob}), (y_3 = \text{:alice}), (y_4 < 30)\\
	\end{split}
\end{equation*}
\normalsize

\begin{algorithm}[t]
	\footnotesize
	\begin{algorithmic}[1]
		\REQUIRE
		$\mathcal{E}$ an enriched entity set $\{E_{e0}, E_{e1}, \dots\}$

		\ENSURE
		$\mathcal{Q}$ a set of query rules, \ie, the DB-query with one or more rules

		\STATE  $\mathcal{Q} := \emptyset$ 
		\FORALL{enriched entity $E$ in $\mathcal{E}$ \label{fdfd}}
			\IF{$E$ is $E_{e0}$ \label{algoL:solar}}
				\STATE $\{ (eval, dbTval, lTval, opval) \} := E$
				\STATE $A_0:= \buildAtom(dbTval, x)$ \COMMENT{Build the first atom for the query's body}
				\STATE $\mathcal{Q} := \{ q(x) \leftarrow A_0 \}$
			\ELSE
				\STATE $Parts := \emptyset$  \COMMENT {Set of list of atoms. Each $l \in Parts$  is a list of atoms\\whose conjunction should be added to query's body.}
				\STATE $E' := \emptyset$ \COMMENT{Set  storing $E$'s tuples already treated}
				\STATE \COMMENT{Treatment of entities  enriched with a  context}
				\FORALL{tuple $t = (eval, dbTval, \text{"Context"}, opval)$ in $E$ \label{algoL:startEC}}
					\STATE Let $t' \in E$,  s.t $t' = (eval, dbTval', lTval', =)$ and $lTval' \neq \text{"Context"}$
					\STATE $y:= GetNewVar()$
					\STATE {$Parts := Parts \cup\{\langle  \buildAtom(dbTval', y), \buildAtom(dbTval, y),$ \label{algoL:stECinParts}}
					\hskip\algorithmicindent $\buildAtomOP(eval, y, opval)\rangle\}$
					\STATE {$E' := E' \cup \{t', t\}$ \label{algoL:endEC}}
				\ENDFOR
				\STATE \COMMENT Treatment of enriched  entities without tuples where\\$lTval=\text{"Context"}$
				\FORALL{tuple $t = (eval, dbTval, lTval, opval)$ in ($E \setminus E'$) \label{algoL:startNonEC}}
						\STATE $y:= GetNewVar()$
						\STATE {$Parts := Parts \cup\{\langle  \buildAtom(dbTval, y),  \buildAtomOP(eval, y, opval)\rangle\}$
						\label{algoL:endNonEC}}
				\ENDFOR
				\STATE $ $
				\STATE $\mathcal{Q}' := \emptyset$
				\FORALL{query $q \in \mathcal{Q}$}
					\FORALL{list $l \in Parts$}
						\STATE {$q' = BuildNewQuery(q, l)$ \label{algoL:l2q}}
						\STATE $\mathcal{Q}' := \mathcal{Q}' \cup \{q'\}$
					\ENDFOR
				\ENDFOR
				\STATE $\mathcal{Q} := \mathcal{Q}'$
			\ENDIF
		\ENDFOR
		\RETURN $\mathcal{Q}$
	\end{algorithmic}
	\caption{EntitiesToQueries}
	\label{algo:EntitiesToQueries}
\end{algorithm}

However, if we consider $E_ {enew}$ of Example~\ref{ex:or}, 
Algorithm~\ref{algo:EntitiesToQueries} (lines~\ref{algoL:startEC} to~\ref{algoL:endEC}) produces
two lists from the same entity, namely,
$l_1 = \langle Person (y_2),$ $ writtenBy(x, y_2),$  $(y_2= \text{:bob})   \rangle$
and 
$l_2 = \langle Person (y_3),$ $ writtenBy(x, y_3),$  $(y_3= \text{:alice})   \rangle$.
Then, on line~\ref{algoL:l2q}, each list is considered separately and the query $q$ is replaced by two new queries.
At the end, $\mathcal{Q}$ returns a DB-query composed by two rules:

\vspace{-1em}
\small
\begin{equation*}
	\begin{split}
		q(x) \leftarrow & Book(x),  hasTitle(x,y_1), Person(y_2),  (y_1= "Princ. ~of ~Medicine"),\\
		&  writtenBy(x, y_2), hasPrice(x, y_4), (y_2 = \text{:bob}), (y_4 < 30)\\
		q(x) \leftarrow & Book(x),  hasTitle(x,y_1), Person(y_3),  (y_1= "Princ. ~of ~Medicine"),\\
		&  writtenBy(x, y_3), hasPrice(x, y_4), (y_3 = \text{:alice}), (y_4 < 30)\\
	\end{split}
\end{equation*}
\normalsize

Finally, consider
$E_{enew2} = \{(\text{:bob},\ Person,\ Person,\ =),\ (\text{:bob},\ Author, $ \linebreak $Context,\ =),$
$(\text{:bob},\ Editor,\ Context,\ =)\}$.
The resulting lists on line~\ref{algoL:stECinParts} are
$l1=  \langle Person (y_2),$ $ writtenBy(x, y_2),$  $(y_2= \text{:bob})   \rangle$
and 
$l_2 = \langle Person (y_3),$ $ editedBy(x, $ $y_3),$  $(y_3= \text{:bob})   \rangle$
and  $\mathcal{Q}$ also returns a DB-query composed by two rules.
Here we are looking for \textit{books edited or written by Bob}.

Thus, queries can be directly generated from enriched entities. 
Currently we only deal with  conjunctive  queries -- easily translated to SQL or SPARQL.
\vspace{-.5em}

\section{Implementation and Experimental Results}
\label{sec:implementation}
\vspace{-.5em}
In order to validate our system, we implemented it in the form of a pipeline which allows us to divide the separate stages and explore various combinations.
For lexicon-based entity extraction, Apache SolR~\cite{ApacheSolr} is used with its text tagger, an inverted index and $n$-gram algorithm~\cite{kimFastStringMatching1994}.
It allows lexemes detection even with typographic errors.
We also use a combination of hand-written grammars together with a Facebook project called Duckling~\cite{duck} 
which provides powerful tools for extracting  entities such as numbers or dates.
Each extraction step is performed independently and simple entities are defined by taking into account all different methods.
In particular, if several approaches identify entities in the same place (but not necessarily with the same bounds), we keep only the  entity resulting from the union of the overlapping entities to represent the ambiguity.
To implement this pipeline and link each step, we use the RASA NLU framework~\cite{bocklischRasaOpenSource2017} in combination with SpaCy for the parsing phase.
Notice that  the pre-processing part is based on generic grammars and lexicons.
Some lexicons are generated automatically by considering values in the RDF database (\eg, first and last name for $Person$).
Hand-written lexicons such as those for operators and contexts concerning dates (\eg, application, archive, creation, expiration) or persons (\eg, author, signatory) are also used.
Partial matches are managed using multiple lexemes when possible (\eg \textit{create by}, \textit{create with}, \textit{create}, \dots).

We conduct our preliminary experiments  on an RDF database concerning medical publications.
The database has 66 classes (possible candidates for a query solar-class) with a total of 29327 class instances.
In our tests, about $10$ classes have been used as solar-classes.
These tests have considered  entity extraction and enrichment phases.
Ambiguity has not been tested and thus we only take into account one value per entity.
The evaluation is done by analysing the obtained enriched entities.
Table~\ref{tab:results} shows the results of our experiments on a set of $113$  NL-queries (varying number of \textit{and} and \textit{or}).
It summarizes the precision, recall, f1-score, and the weighted mean (weighted by support) obtained for each \dbtype\ on the NL-queries set.
As our system is implemented as a pipeline, we intend to perform tests step by step, in order to  identify the impact of each step.



\begin{wrapfigure}[12]{R}{0.5 \textwidth}
    \vspace{-2em}
    \scriptsize
    \centering
    \begin{tabular}{l | c c c c}
                   $DBType$       & precision & recall & f1-score & support \\
        \hline
        solar\_class      & 1.00      & 0.62   & 0.77     & 82      \\
        application\_date & 1.00      & 0.62   & 0.76     & 13      \\
        archive\_date     & 0.50      & 0.67   & 0.57     & 3       \\
        creation\_date    & 0.60      & 0.60   & 0.60     & 5       \\
        expiration\_date  & 1.00      & 0.50   & 0.67     & 2       \\
        customers         & 1.00      & 0.60   & 0.75     & 5       \\
        department        & 1.00      & 0.11   & 0.20     & 9       \\
        sector            & 1.00      & 0.95   & 0.98     & 21      \\
        doc\_author       & 0.77      & 0.63   & 0.70     & 38      \\
        doc\_signatory    & 0.90      & 0.86   & 0.88     & 21      \\
        doc\_status       & 1.00      & 0.50   & 0.67     & 6       \\
        doc\_unit         & 1.00      & 0.27   & 0.43     & 11      \\
        \dots             & \dots     & \dots  & \dots    & \dots   \\
        \hline
        \textbf{Weighted avg.}     & 0.86      & 0.59   & 0.67     & 295     \\
    \end{tabular}
    \caption{Results on enriched entities}
    \label{tab:results}
    \vspace{-.8em}
\end{wrapfigure}

Our precision is good, indicating that most of our detected entities are  the expected ones.
This is clearly a consequence of the effective use of lexicons and grammars.
For recall, our results are not bad, but weaker than our precision, indicating that some entities are not detected.
Lower precision on some \dbtype\ like \textit{creation\_date} or \textit{doc\_author} is partially explained because ambiguity is not taken into account in our experiment, but  entities giving rise to these types are enriched with a similar context  and associated with both types. 
A similar issue occurs with the recall for \textit{department} and \textit{doc\_unit}. In our test database they are semantically close.
So, we have significant overlap on the two lexicons (a lot of lexemes are shared, adding ambiguity to the type).
Our current work consists in improving lexicons for context detection, in particular those generated automatically.
\vspace{-.5em}
 
%

\section{Related Work and Concluding Remarks}
\label{sec:rw}
\vspace{-.5em}
Recently,  NLI has been widely discussed in the literature.
Some work focuses on augmenting the expression power of queries while others on domain-independence.
For instance, in~\cite{zhengQuestionAnsweringKnowledge2018} authors propose the use of binary templates rather than semantic parses to better understand complex queries while~\cite{wangCrossdomainNaturalLanguage2019}  proposes a cross-domain NLI with  a general propose question tagging strategy.
Several work (such as in~\cite{ZHWYHZ14,AKM15,FZE14}) consider RDF question/answering (QA).
Aggregate queries are considered in~\cite{HDYY18}: the authors propose a method to automatically identify the aggregation  and transform it into a SPARQL aggregate statement.
Methods used vary a lot.  In~ \cite{steinmetzNaturalLanguageQuestions2019,zafarFormalQueryGeneration2018,zhengQuestionAnsweringKnowledge2018} authors base their approach on NLP techniques with entity extraction and grammars, while in \cite{utamaEndtoendNeuralNatural2018,wangCrossdomainNaturalLanguage2019}  they use  neural networks.

\vspace{0.1cm}
The paper presents a method where enriched entities allow us to translate NL-queries into DB-queries.
The use of a logical paradigm to deal with databases shows that our method can be adapted to different data models.
Our approach is divided into a domain-dependent pre-processing and domain-independent query generation phases. 
The first step, responsible for building lexicons,  grammar-tools  and ontology mappings, also sets up general propose tools which can be considered as domain-independent (\eg, grammars for date recognition).
The second step of our method can be applied on any domain, provided the ontology mappings are set up.
This  division allows us to deal with possible extensions  and improvements separately. 
We are currently considering extensions of Algorithm~\ref{algo:EntitiesToQueries}  in order to deal with queries on more than one solar-class 
or aggregate queries.
We also plan to extend entity extraction by including alternative approaches such as machine learning to complete grammars (\eg~for title identification).

\bibliographystyle{splncs04}
\bibliography{refs}

\end{document}